\documentclass{article}

\usepackage[final]{nips_2016}

\usepackage[utf8]{inputenc} % allow utf-8 input
\usepackage[T1]{fontenc}    % use 8-bit T1 fonts
\usepackage{url}            % simple URL typesetting
\usepackage{booktabs}       % professional-quality tables
\usepackage{amsfonts}       % blackboard math symbols
\usepackage{nicefrac}       % compact symbols for 1/2, etc.
\usepackage{microtype}      % microtypography
\usepackage{amsmath}
\usepackage{subfigure}
\usepackage{graphicx}

\title{Learning Interpretability for Visualizations using Adapted Cox Models through a User Experiment}

\author{
  Adrien Bibal\\
	PReCISE Research Center\\
  	Faculty of Computer Science\\
  	University of Namur\\
  	Namur, 5000 - Belgium\\
  \texttt{adrien.bibal@unamur.be} \\
  \And
  Benoît Frénay \\
	PReCISE Research Center\\
	Faculty of Computer Science \\
	University of Namur\\
	Namur, 5000 - Belgium\\
	  \texttt{benoit.frenay@unamur.be} \\
}

\begin{document}

\maketitle

\begin{abstract}
	In order to be useful, visualizations need to be interpretable. This paper uses a user-based approach to combine and assess quality measures in order to better model user preferences. Results show that cluster separability measures are outperformed by a neighborhood conservation measure, even though the former are usually considered as intuitively representative of user motives. Moreover, combining measures, as opposed to using a single measure, further improves prediction performances.
\end{abstract}

\vspace{-.1cm}
\section{Introduction}
\vspace{-.1cm}
Measuring interpretability is a major concern in machine learning. Along with other classical performance measures such as accuracy, interpretability defines the limit between black-box and white-box models \citep{Ruping2006, bibal2016}. Interpretable models allow one to understand how inputs are linked to the output. This paper focuses on visualizations that map high-dimensional data to a 2D projection. In this context, interpretability refers to the ability of a user to understand how a particular visualization model projects data.  When a user chooses a particular visualization, he or she implicitly states that he or she understands how the points are presented, i.e. how the model works. Interpretability is then defined through user preferences and no a priori definition is assumed. 

Following \cite{Freitas2014} and others, \cite{bibal2016} highlights two ways to measure interpretability: through heuristics and user-based surveys. Tailored quality measures for visualizations are examples of the heuristics approach. Surveys can be used to qualitatively define the understandability of a visualization by asking for user feedback. Both approaches are complementary, but only a few works (e.g. \cite{sedlmair2015}) attempt to mix them to assess the relevance of several quality metrics for visualization. This paper bridges this gap through a user-based experiment that uses meta-learning to combine several measures of visualization interpretability.

Section \ref{measures} presents some visualization quality measures that are used during meta-learning. Section \ref{cox} introduces a family of white-box meta-models to find a score of interpretability.  Then, Section \ref{experiment} describes the user experiment that is used to model interpretability from user preferences. Finally, Section 5 discusses the experimental results and Section 6 concludes the paper.
\vspace{-.1cm}
\section{Quality Measures of Visualizations\label{measures}}
\vspace{-.1cm}
One can consider two types of quality measures for visualizations: one type uses only the data after projection and the other compares the points before and after projection. Typical measures of the first type focus on the separability of clusters in the visualization. \cite{sedlmair2015} reviewed, evaluated and sorted such measures in terms of algorithmic similarity and agreement with human judgments. They confirmed the top position of distance consistency (DSC) as one of the best measures \citep{sedlmair2015}. Let $\mathcal{P}$ be the set of points of the projection, $\mathcal{C}$ the set of classes and $centroid(c)$ the centroid of class $c$, then \citep{sips2009}:
\begin{equation*}
\text{DSC} = |\{x \in \mathcal{P} : (\exists c \in \mathcal{C} : c \not= c_{x} \land \text{dist}(centroid(c), x) < \text{dist}(centroid(c_{x}), x))\}|\,/\,|\mathcal{C}|.
\end{equation*}

Two other top measures in \cite{sedlmair2015} are the hypothesis margin (HM) and the average between-within (ABW). HM computes the average difference between the distance of each point $x$ from its closest neighbor of another class and its closest neighbor of the same class \citep{gilad2004}. ABW \citep{sedlmair2015, lewis2012} computes the ratio of the average distance between points of different clusters and the average distance within clusters.

In order to compare visualization algorithms, \cite{lee2015} propose a measure of the second type modeling neighborhood preservation. Their measure, $\text{NH}_{\text{AUC}}$, can be defined as follows. Let $N$ be the number of points in the dataset, $K$ the number of neighbors, $v^K_i$ the $K$ nearest neighbors of the $i$th point in the original dataset and $n^K_i$ the $K$ nearest neighbors of the $i$th point in the projection,
\begin{equation*}
\text{Q}_{\text{NX}}(K) = \left(\sum\limits_{i=1}^N |v^K_i \cap n^K_i|\right)/(KN)
\end{equation*}
measures the average preservation of neighborhoods of size $K$. \cite{lee2015} then use the area under the $\text{Q}_{\text{NX}}(K)$ curve for different neighborhood sizes in order to compute $\text{NH}_{\text{AUC}}$.
\vspace{-.1cm}
\section{Meta-Learning with Adapted Cox Models\label{cox}}
\vspace{-.1cm}
The main goal of this paper is to evaluate whether combining state-of-the-art measures of different types improve the modeling of human judgment. To asses this, we set up an experiment asking users to express preferences between visualizations shown in pairs (see section~\ref{experiment} for more details) and then used these preferences to determine an interpretability score. Since our dataset is composed of preferences between visualizations, our learning problem is rooted in preference learning. For this kind of problem, an order must be learned based on preferences \citep{furnkranz2011}. Our dataset consists of a set of visualizations $\mathcal{V}$ and a set of user-given preferences $v_i \succ v_j$ expressing that $v_i$ is preferred over $v_j$ for some pairs of visualization $v_i, v_j \in \mathcal{V}$.

The preference learning algorithm considered for modeling user preferences must be interpretable, such as with a logistic regression \citep{arias2008}, so that knowledge about the measures used as meta-features can be gained. To solve this problem, we consider a well-known interpretable model used in survival analysis, the Cox model \citep{cox1972, Branders2015}. We adapted the Cox model to fit our preference learning problem. Indeed, in the case of pairwise comparisons of objects, the partial likelihood of a Cox model can be adapted as follows:
\begin{equation*}
\text{Cox}_{\text{pref}}(\beta) = \prod\nolimits_{v_i \succ v_j} \Bigl[\frac{exp(\beta^T v_i)}{exp(\beta^T v_i) + exp(\beta^T v_j)}\Bigr] = \prod\nolimits_{v_i \succ v_j} \Bigl[\frac{1}{1 + exp(-\beta^T (v_i - v_j))}\Bigr].
\end{equation*}
This adapted Cox model learns a preference score using measures presented in section \ref{measures} as features of visualizations $v_i$ and $v_j$. This regression differs from a true logistic regression in that there is no intercept term. The term $\beta^Tv_i$ can be interpreted as an understandability score for visualization $v_i$.
\vspace{-.1cm}
\section{User-Based Experiment\label{experiment}}
\vspace{-.1cm}
As mentioned in section \ref{cox}, an experiment was set up to collect preferences from users. Visualizations presented to users were generated from the dataset MNIST with various numbers of classes (from 2 to 6) using t-SNE \citep{maaten2008} with various perplexities between $5$ and the dataset size in a logarithmic scale. Each user was interviewed after the experiment to discuss his or her strategies for choosing between visualizations. We then used this information to better understand cases where Cox$_\text{pref}$ models were not in agreement with user preferences.

The population of our experiment consisted of 40 first-year university students. They were instructed to select, from two  displayed visualizations, the one for which they best understood  ``how the computer had positioned the numbers''. In addition to these two options, they could also select ``no preference'', in which case the comparison was not used for learning. Successive comparisons were assumed to be independent, meaning that no psychological learning bias was assumed to be involved. 

A total of 3294 preferences was collected. Because each user may have a different strategy while choosing visualizations, they were grouped into batches per user. For a given user, a random subset of his or her preferences was selected, with the total number of preferences being the same for all users. Thanks to this subsampling, all users had the same weight when modeling the overall strategy. The number of preferences per user was set at 30, which let aside 10 users that provided less than 30 preferences; our dataset was composed of 900 preferences. 1000 user permutations were performed. For each permutation, 2/3 of the users were used for training the Cox$_\text{pref}$ model and 1/3 for testing. The performance measure was the percentage of agreement between users and the model. We used the same performance measure to individually compare the visualization measures used as meta-features.
\vspace{-.1cm}
\section{Discussion\label{results}}
\vspace{-.1cm}
In addition to the two types of measures presented in section \ref{measures}, the number of classes was also considered for meta-learning \citep{garcia2016}. In the case of a tie (i.e same number of classes), one of the visualization was chosen randomly. Table \ref{pref} shows the means and standard deviations computed on the 1000 permutations and table \ref{winlose} presents the percentage of win against other measures. Measure $m^p_i$ wins against measure $m^p_j$ if $m_i$ has better performances than $m_j$ for the permutation $p$.

\begin{table}[t]
	\caption{Average percentage of agreement with user preferences and 95\% confidence interval thereof.}
	\label{pref}
	\centering
	\begin{tabular}{c|c|c|c|c|c}
		number of classes & ABW & HM & DSC & $\text{NH}_{\text{AUC}}$ & Cox$_\text{pref}$ \\
		\hline
		63.6\% $\pm$ 0.1 & 65.6\% $\pm$ 0.1 & 67\% $\pm$ 0.2 & 68.5\% $\pm$ 0.2 & 71.5\% $\pm$ 0.1 & 76.4\% $\pm$ 0.2\\
	\end{tabular}
\end{table}

\begin{table}[t]
	\caption{Percentage of wins for every pairwise comparison between the five quality measures.}
	\label{winlose}
	\centering
	\begin{tabular}{c|c|c|c|c|c|c}
		& number of classes & ABW & HM & DSC & $\text{NH}_{\text{AUC}}$ & Cox$_\text{pref}$ \\
		\hline
		ABW                      & 84.5\%  &     &    &   &    &    \\
		HM                       & 88.3\%  & 67\%   &     &     &    &    \\
		DSC                      & 97.5\%  & 89.6\% & 70\%   &      &  &    \\
		$\text{NH}_{\text{AUC}}$ & 100\%   & 99.3\% & 98.2\% & 87.1\%  &      &  \\
		$Cox_\text{pref}$        & 100\%   & 100\%  & 100\%  & 100\%   & 99.3\%  &    \\
	\end{tabular}
\end{table}

Among the measures of the first type discussed in section \ref{measures}, DSC performs well in its group but is beaten by $\text{NH}_{\text{AUC}}$, the measure of the second type. Interestingly, $\text{NH}_{\text{AUC}}$ obtains very good results despite the fact that it does not directly apply the well-known user-strategy of cluster separability \citep{sedlmair2015}, a strategy that was confirmed during the interviews. Indeed, measures of the second type use the original high-dimensional data in their computation, which is not possible for a human. In both table \ref{pref} and \ref{winlose}, the Cox$_\text{pref}$ model outperforms individual measures. Similar results were observed using all 3129 preferences from the same 30 users.

\begin{figure}[h]
	\centering
	\subfigure[]{\includegraphics[width=.32\textwidth]{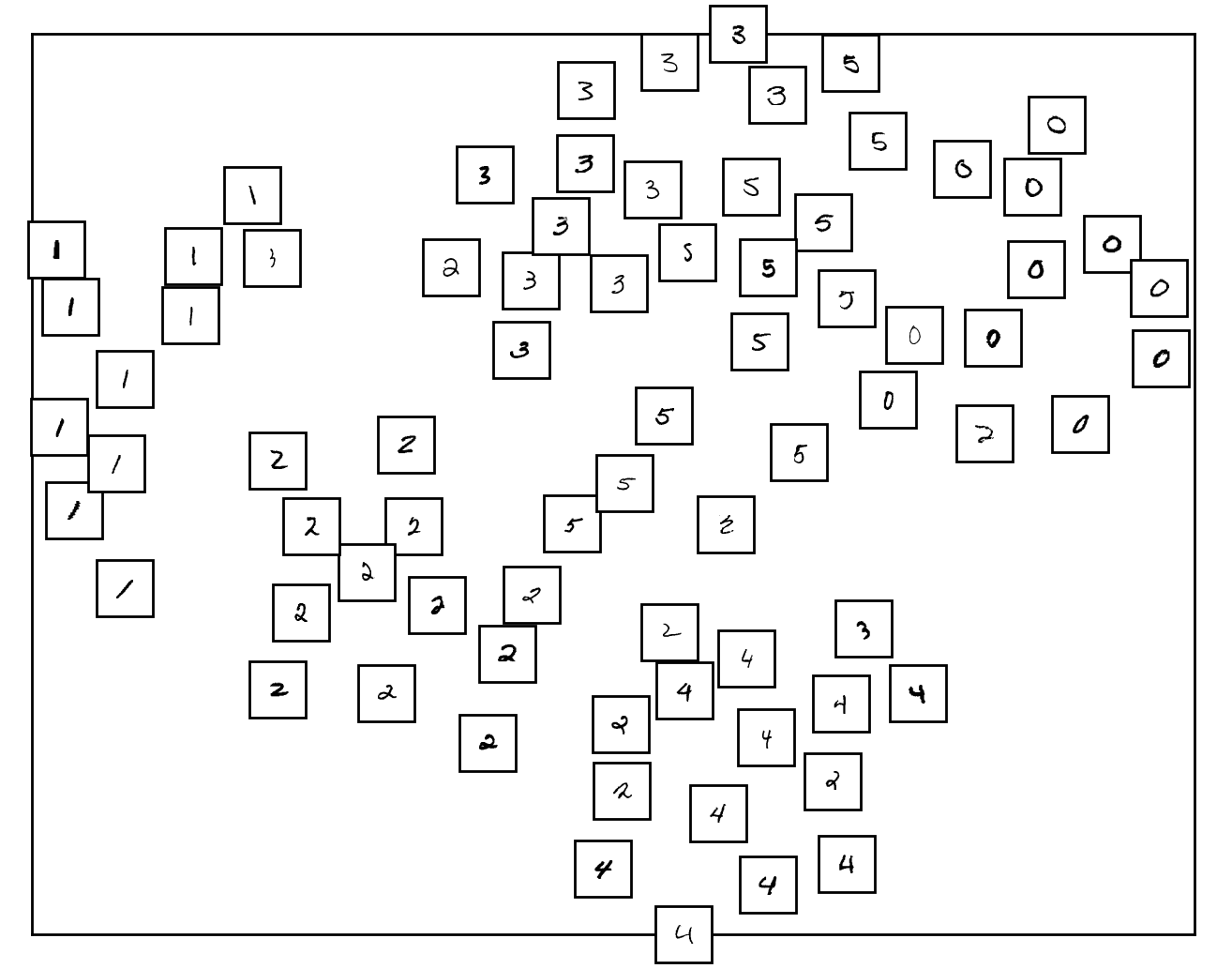}}
	\subfigure[]{\includegraphics[width=.32\textwidth]{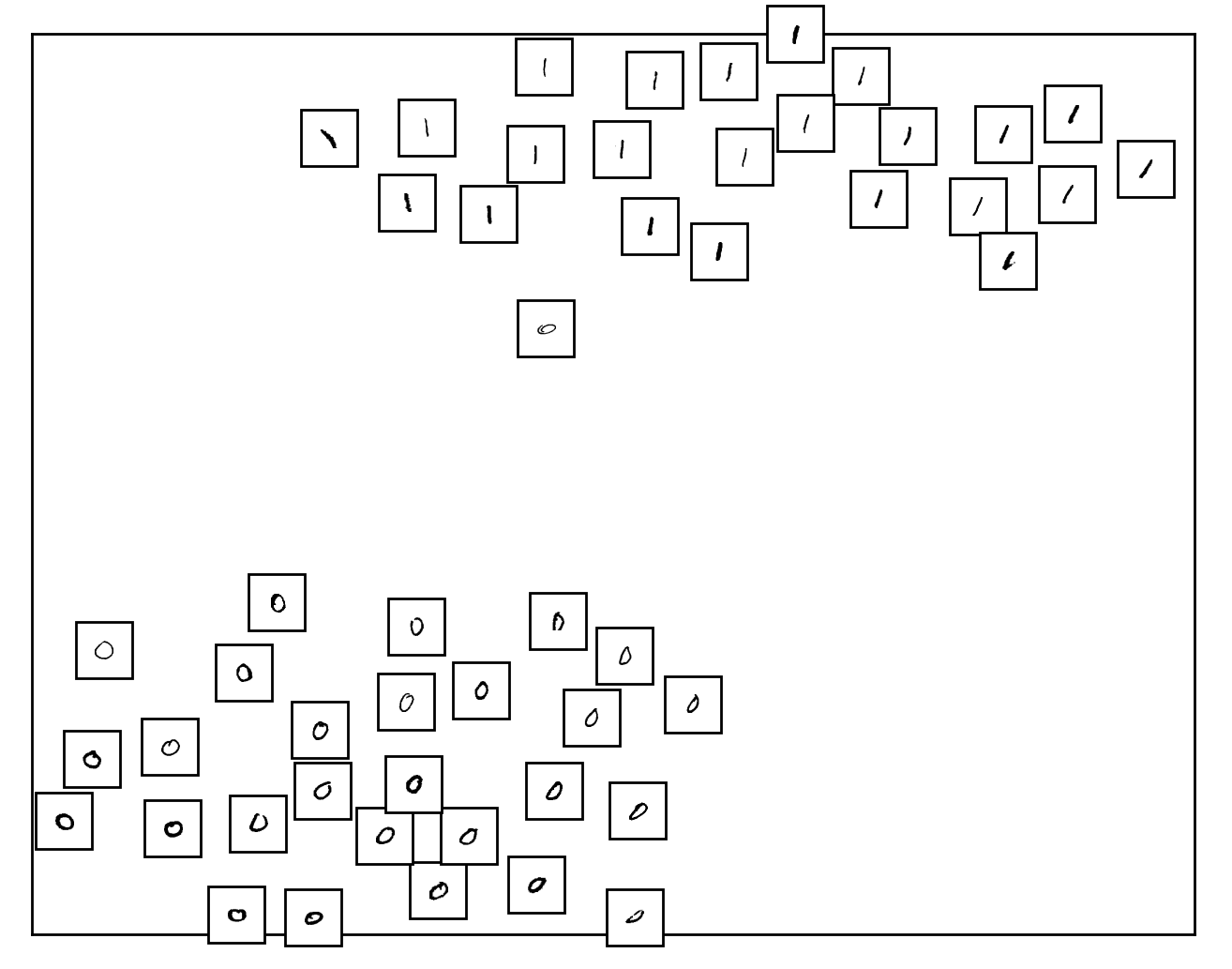}}
	\subfigure[]{\includegraphics[width=.32\textwidth]{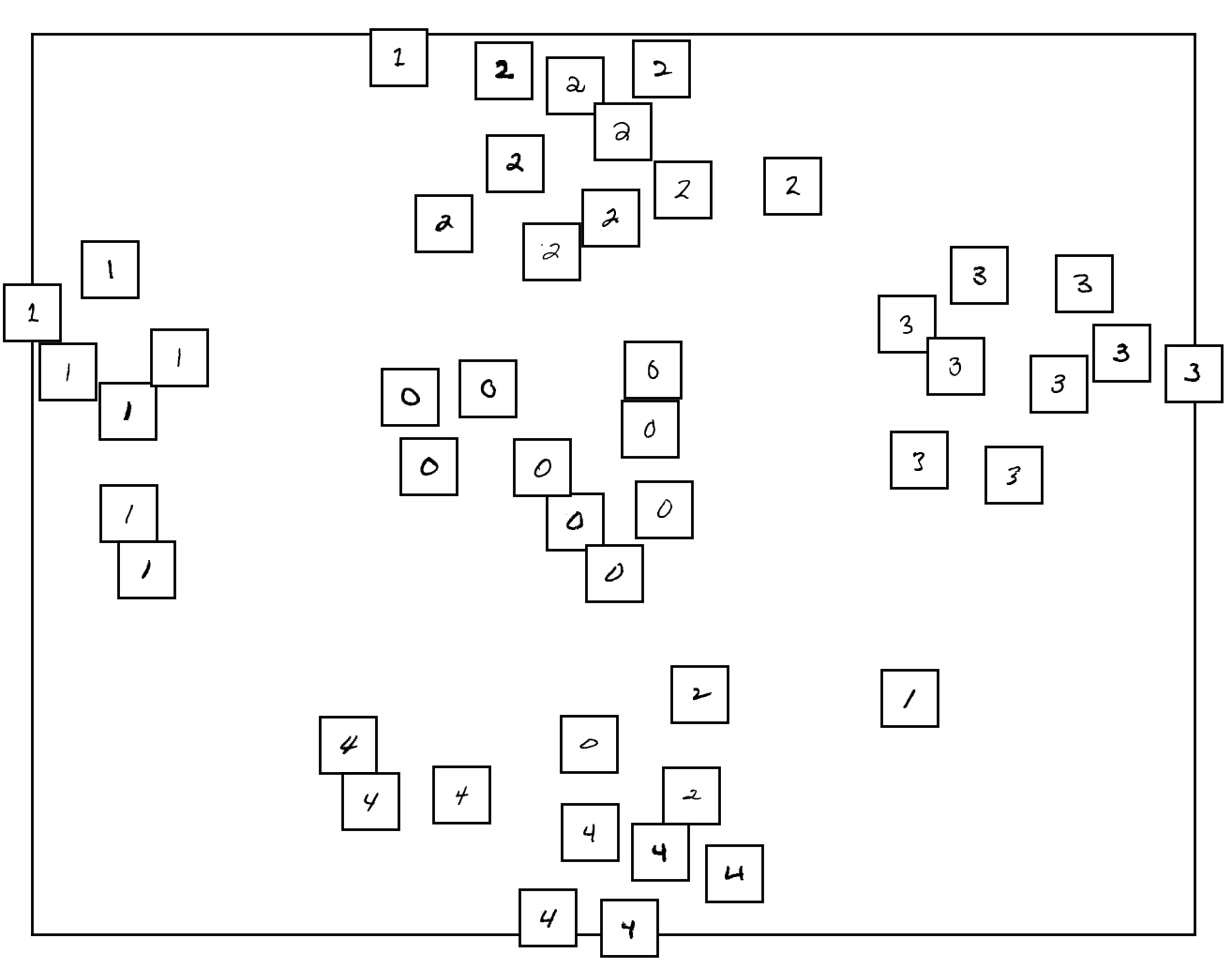}}

	\caption{Examples of disagreement between users and Cox$_\text{pref}$. Among visualizations (a) and (b), Cox$_\text{pref}$ prefers (b) where 0s and 1s are clearly separated, whereas the user preferred (a). Visualization (c) shows an example of semantic bias: two users reported that they preferred (c) when there is a tie because it looks like a clock (1s on the left, 2s at the top, 3s on the right and 4s at the bottom).}
	\label{horloge}
\end{figure}

In order to understand why the Cox$_\text{pref}$ models fail in 23.6\% of the cases on average, we checked judgment errors from Cox$_\text{pref}$ by referring to what users said during the interviews. We could observe that involving users open the opportunity for mistakes or unusual behaviors, as we can see in figure \ref{horloge}. Furthermore, in a few cases, when the user has no preference but distinguishes a semantic pattern that makes sense for him or her in the visualization, he or she tends to choose it (see figure \ref{horloge}).

In order to assess the importance of each visualization measure in the score of Cox$_\text{pref}$, we varied the L1 penalization to enforce sparsity. $\text{NH}_{\text{AUC}}$ is selected first. Then ABW is added with an improvement of roughly 3.5\%. The number of classes is added as a third measure, which improves the model by roughly 1.5\%. Other additional measures only offer a minor improvement.
\vspace{-.1cm}
\section{Conclusion\label{conclusion}}
\vspace{-.1cm}
Using an adapted Cox model to handle the task of preference learning, we observed the modeling power of a measure taking into account elements that a human being cannot handle, such as $\text{NH}_{\text{AUC}}$. Furthermore, we confirmed the position of DSC as leader of its category. Finally, we showed that using a white-box model to aggregate state-of-the-art measures can improve the prediction of human judgment using information of measures from different families. Further work needs to confirm the results obtained with t-SNE for MNIST on a wide range of datasets and visualization schemes.

\vspace{-.1cm}
\section*{Acknowledgments}
\vspace{-.1cm}
We are grateful to Prof. Bruno Dumas for his help for the design of the experiment involving users. We also thanks Dr. Samuel Branders for fruitful discussions and sharing resources on Cox models.

\bibliographystyle{apalike}
\bibliography{biblio.bib}

\end{document}